\documentclass[10pt,twocolumn,letterpaper]{article}

\usepackage{cvpr}

\usepackage{amsfonts}
\usepackage{amsmath}
\usepackage{array}
\usepackage{booktabs}
\usepackage{multirow}
\usepackage{amssymb}
\usepackage{pifont}
\usepackage{xcolor}
\usepackage{lipsum}
\usepackage{xspace}
\usepackage{mwe}
\usepackage{colortbl}

\newcommand{\cmark}{\ding{51}}  
\newcommand{\xmark}{\ding{55}}  

\definecolor{changecolor1}{RGB}{255,0,0}

\definecolor{changecolor2}{RGB}{0,0,255}

\newcommand{\ours}{Pose-dIVE\xspace}
\newcommand{\hlrow}{\rowcolor{black!6}}

\newcommand{\paragrapht}[1]{\vspace{-16pt}\paragraph{#1}}

\definecolor{cvprblue}{rgb}{0.21,0.49,0.74}
\usepackage[pagebackref,breaklinks,colorlinks,allcolors=cvprblue]{hyperref}

\def\paperID{} 
\def\confName{CVPR}
\def\confYear{2026}

\title{Pose-dIVE: Pose-Diversified Augmentation for Person Re-Identification}

\author{In\`es Hyeonsu Kim$^{1*}$\qquad
Woojeong Jin$^{1*}$\qquad
Soowon Son$^1$\qquad
Junyoung Seo$^1$\qquad
Seokju Cho$^1$\qquad \\
JeongYeol Baek$^2$\qquad
Byeongwon Lee$^2$\qquad
JoungBin Lee$^1$\qquad
Seungryong Kim$^{1\dagger}$\qquad\\[0.5em]
KAIST AI$^1$ \qquad SK Telecom$^2$\\
}

\begin{document}
\maketitle

\def\thefootnote{*}\footnotetext{Equal contributions.}\def\thefootnote{\arabic{footnote}}
\def\thefootnote{$\dagger$}\footnotetext{Corresponding author.}\def\thefootnote{\arabic{footnote}}

\begin{abstract}
Person re-identification (Re-ID) often faces challenges due to variations in human poses and camera viewpoints, which significantly affect the appearance of individuals across images. Existing datasets frequently lack diversity and scalability in these aspects, hindering the generalization of Re-ID models to new camera systems or environments. To overcome this, we propose \textbf{\ours}, a novel data augmentation approach that incorporates sparse and underrepresented human pose and camera viewpoint examples into the training data, addressing the limited diversity in the original training data distribution. Our objective is to augment the training dataset to enable existing Re-ID models to learn features unbiased by human pose and camera viewpoint variations. By conditioning the diffusion model on both the human pose and camera viewpoint through the SMPL model, our framework generates augmented training data with diverse human poses and camera viewpoints. Experimental results demonstrate the effectiveness of our method in addressing human pose bias and enhancing the generalizability of Re-ID models compared to other data augmentation-based Re-ID approaches. 
Our project page is available at: \href{https://cvlab-kaist.github.io/Pose-dIVE}{\texttt{https://cvlab-kaist.github.io/Pose-dIVE}}.
\end{abstract}
    
\section{Introduction}
\label{sec:intro}

\begin{figure}[t]
    \centering
    \includegraphics[width=\linewidth]{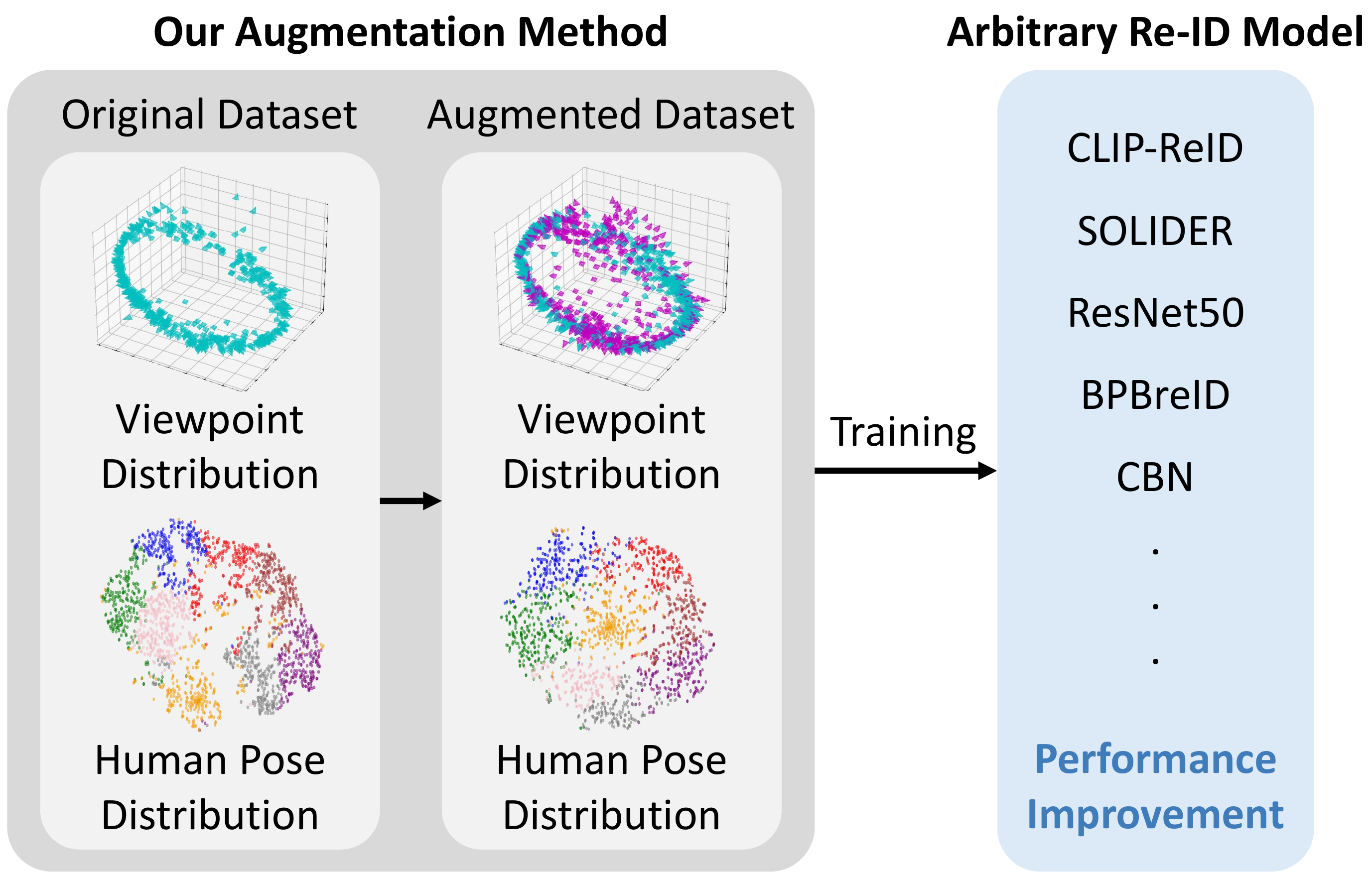}
    \vspace{-15pt}
    \caption{\textbf{\ours} diversifies the viewpoint and human pose of Re-ID datasets to help generalize and improve the performance of arbitrary Re-ID models. On a curated real-world test dataset, the model~\cite{LU-unsuper} trained with the \ours augmented dataset achieves a \textbf{+13.6 mAP} and \textbf{+11.0 R1} improvement (refer Table~\ref{tab:non_ped_evaluation}).
    }
    \label{fig:teaser}
    \vspace{-15pt}
\end{figure}

\begin{figure*}[t]
\centering
\includegraphics[width=1.0\textwidth]{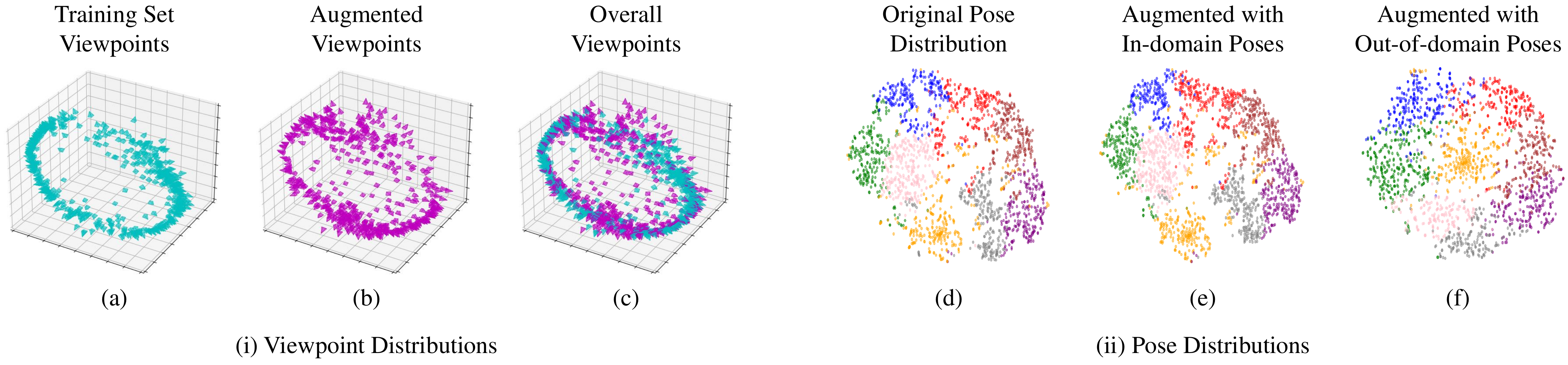}
\vspace{-20pt}
\caption{\textbf{Visualization of the effect of viewpoint and human pose augmentation.} We compare visualizations of camera viewpoint and human pose distributions for the Market-1501~\cite{Market1501}. The left figures (i) display the camera viewpoint distribution derived from SMPL, while the right figures (ii) illustrate the pose distribution. In (i), from left to right, we show the viewpoint distributions of the training dataset, the augmented dataset, and the combination of both. Similarly, in (ii), from left to right, we present t-SNE~\cite{van2008visualizing} visualizations of the human pose distributions, showing poses from the training dataset, followed by augmented poses sourced from outside the dataset. These visualizations demonstrate that \ours successfully diversifies both viewpoint and human pose distributions.
}
\label{fig:multi-cost}\vspace{-10pt}
\end{figure*}
Person re-identification (Re-ID) is widely utilized in modern surveillance systems to track and recognize individuals across multiple camera networks~\cite{wang2013intelligent, zheng2017sift, chen2018person, zheng2016person}. Despite significant advancements in Re-ID methodologies~\cite{zhu2024seas, yang2024pedestrian, SOLIDER, CLIP-reID}, there remains a notable gap between performance under controlled training conditions and effectiveness in real-world scenarios.

Two particularly challenging factors limiting generalization are changes in human pose~\cite{cho2016improving, sarfraz2018pose, zhao2017spindle} and variations in camera viewpoint~\cite{bak2014improving, karanam2015person}. Although an individual's identity remains constant, differences in pose or camera angle can significantly alter their visual appearance. Thus, robust Re-ID models must capture identity-defining features despite these spatial variations. However, current Re-ID datasets often lack sufficient diversity in poses and viewpoints. Typically, these datasets feature only limited walking or standing poses~\cite{zheng2016mars} and employ just two or three camera viewpoints per identity~\cite{Market1501, CUHK03-1, MSMT17}. Such restricted diversity produces unimodal samples that hinder models from learning robust identity representations. Furthermore, testing sets that lack diverse scenarios may not accurately reflect real-world complexities.

Collecting and annotating richer and more varied datasets could mitigate these issues. Nevertheless, privacy concerns and the high cost associated with large-scale, multi-view camera installations~\cite{liu2014semi, zhao2013unsupervised, LU-unsuper} further complicate the collection of richer, more varied datasets. As a result, pose invariant feature learning~\cite{liu2023learning, karanam2015person, karmakar2021pose, ge2018fdgan} and data augmentation~\cite{qian2018pngan, zhong2020random, mclaughlin2015data, zheng2017unlabeled, chen2022learning, chen2021joint} have become critical approaches in addressing this limitation. Existing augmentation techniques primarily exploit the limited range of poses~\cite{liu2018pose, ge2018fdgan} and viewpoints~\cite{chen2022learning, chen2021joint} already present in current datasets. Additionally, these methods usually restrict viewpoint augmentation to horizontal rotations and neglect elevation changes~\cite{chen2022learning, chen2021joint}. Moreover, pose and viewpoint have traditionally been treated as separate factors, despite their combined influence on human appearance.

To address these issues, our work proposes an integrated augmentation strategy that jointly considers pose and viewpoint variations while introducing greater variability in both factors, as illustrated in Figure~\ref{fig:multi-cost}. Instead of relying solely on poses from existing Re-ID datasets, we incorporate dynamic poses sourced externally. We simultaneously adjust both azimuth and elevation angles to achieve a broader and more realistic range of camera viewpoints. This comprehensive augmentation strategy enables Re-ID models to better recognize stable identity features across varied appearances. Our experiments validate this approach by demonstrating improved performance even when existing models are trained with these augmented samples.

Our approach utilizes recent large-scale diffusion models~\cite{rombach2022high}, leveraging their ability to encode extensive prior knowledge about human appearances under diverse conditions. By conditioning these models within a dual-branch architecture~\cite{AnimateAnyone} that simultaneously preserves reference identity and integrates SMPL-derived~\cite{loper2015smpl} pose and viewpoint guidance, we generate high-fidelity training samples. This specifically targets distributional gaps found in conventional Re-ID datasets.

To rigorously evaluate model robustness, we test systems trained with our augmented dataset on both real-world data and a customized evaluation set containing poses and viewpoints absent from the training data. This methodology effectively measures how well models generalize to unseen conditions and practical scenarios. As shown in Figure~\ref{fig:teaser}, our experiments confirm that training on datasets augmented by \ours significantly enhances Re-ID performance on standard benchmarks and in more challenging real-world conditions. 

\begin{figure*}[t]
\centering
\includegraphics[width=1\textwidth]{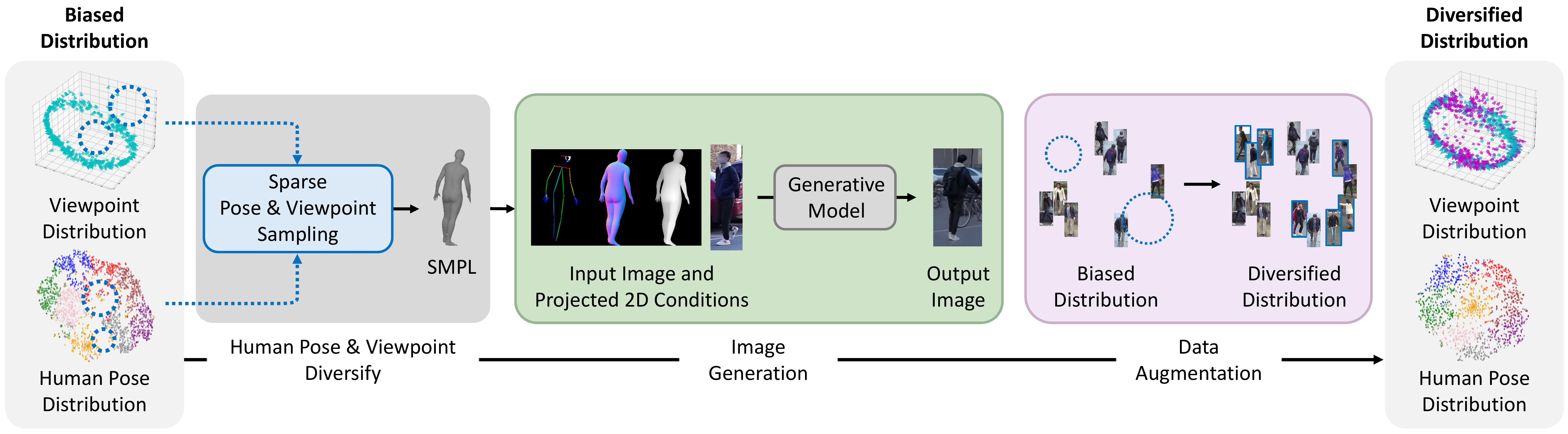}
\vspace{-20pt}
\caption{\textbf{\ours framework.} Upon observing the highly biased viewpoint and human pose distributions in the existing training dataset, we augment the dataset by manipulating SMPL body shapes and feeding the rendered shapes into a generative model to fill in sparsely distributed poses and viewpoints. With this augmented dataset, we can train a Re-ID model that is robust to viewpoint and human pose biases.}
\vspace{-15pt}
\label{fig:motivation}
\end{figure*}
\section{Related Work}

\paragraph{Data augmentation in person re-identification.}
Person re-identification (Re-ID) has advanced significantly with deep learning techniques, enabling robust matching of individuals across non-overlapping camera views ~\cite{SOLIDER, CLIP-reID, BPBreID}. However, a persistent challenge in Re-ID is the scarcity of diverse training datasets, which limits model generalization. This issue has spurred exploration into various strategies, including data augmentation, to enhance the robustness and scalability of Re-ID systems.

Early augmentation methods adopted straightforward image transformations such as random resizing, cropping, and horizontal flipping ~\cite{luo2019strong}, as well as techniques to simulate occlusions~\cite{huang2018adversarially}. These approaches enrich the training data, providing benefits to methods like distance metric learning ~\cite{koestinger2012large, liao2015efficient}, which rely on diverse samples to better embed features of the same identity closely together while separating those of different identities. While effective in increasing data variety, these techniques primarily leverage variations already present in the dataset, leaving gaps in addressing underrepresented aspects such as human pose and camera viewpoint diversity.

\paragrapht{Generative data augmentation in Re-ID.}
More advanced methods have employed Generative Adversarial Networks (GANs) ~\cite{goodfellow2020generative} to synthesize training images. LSRO ~\cite{zheng2017unlabeled} utilizes DCGANs ~\cite{radford2015unsupervised} to generate unlabeled samples, enhancing semi-supervised Re-ID through label smoothing regularization. Similarly, pose-transferrable GANs ~\cite{liu2018pose} augment datasets by transferring poses, though they rely on poses extracted from existing Re-ID data. Other works, such as ~\cite{zheng2019joint}, focus on generating cross-ID images to support joint discriminative and generative learning. These GAN-based approaches have demonstrated performance improvements by expanding the training data. However, their emphasis often lies in utilizing variations already present within the dataset or addressing objectives such as labeling or identity synthesis. In contrast, our work focuses on diversifying the training dataset by uniformly sampling viewpoints and human poses from external datasets.

\paragrapht{3D mesh guidance in Re-ID.}
Several studies have utilized 3D human mesh representations, such as SMPL~\cite{loper2015smpl}, to model human body shape in Re-ID tasks. For instance, GCL ~\cite{chen2021joint} employs a 3D mesh-based view generator that rotates the mesh horizontally to create new viewpoints while preserving the given human pose. This preservation of the original pose contrasts with our objective, as we aim to diversify both poses and viewpoints in the training data. Other efforts, such as 3DInvarReID ~\cite{liu2023learning}, focus on long-term Re-ID and 3D body shape reconstruction, which are distinct tasks from our goal of enhancing training data diversity. Similarly, \citet{chen2022learning} uses a 3D mesh to guide GANs for unsupervised Re-ID, but it relies on poses and viewpoints already available in the Re-ID dataset. Our approach, however, integrates external pose and viewpoint data, setting it apart from these in-domain strategies. 

\section{Method}

\subsection{Overview}

In this paper, we propose a data augmentation strategy designed to address the limitations of existing Re-ID datasets, particularly the restricted range of camera viewpoints and human poses, which is depicted in Figure~\ref{fig:motivation}. In this section, we first explain the augmentation process, detailing how camera viewpoints and human poses are represented during augmentation. Next, we describe how these conditions, along with identity information, are incorporated into the diffusion model, which is designed to accommodate these conditions. The data generated using our approach can be applied to any Re-ID model, enhancing its generalizability.

\subsection{Human Pose and Camera Viewpoint Condition with SMPL}\label{sec:smpl_condition}
A key component of our augmentation strategy is the ability to generate diverse human poses and camera viewpoints in a controlled manner. To accomplish this, we provide specific conditions to the generative model to guide the augmentation process.

Previous works~\cite{ge2018fdgan, PN-GAN, tang2020xinggan} have primarily addressed pose control by providing a human pose skeleton. However, relying solely on the human skeleton has limitations due to missing information: when projected onto 2D images, the human skeleton lacks depth information, posing ambiguity for the model when inferring viewpoint information from the skeleton. For example, if the camera is above a person, the skeleton would appear compressed vertically. Without depth information, the model cannot distinguish whether the camera is positioned above the person or the person is simply short.

In this regard, in addition to the human pose skeleton, we utilize SMPL~\cite{loper2015smpl}, a human body model used for realistic human rendering. SMPL can model intricate human shapes, including complex body articulations in 3D space. From this human model, we can easily extract 2D representations of the 3D human by rendering the model, such as depth maps which implicitly contain camera viewpoint information.

Let us define the SMPL model as a function $\mathcal{M}$ that generates a 3D mesh based on shape parameters $\beta \in \mathbb{R}^{10}$ and pose parameters $\theta \in \mathbb{R}^{72}$:
\begin{equation}
 \{\mathcal{V}, \mathcal{F}\} \leftarrow \mathcal{M}(\beta, \theta),
\end{equation}
where $\mathcal{V} \in \mathbb{R}^{N \times 3}$ represents the set of $N$ vertices in 3D space and $\mathcal{F}\in \mathbb{N}^{F \times 3}$ represents the set of $F$ triangular faces, each defined by three vertex indices from $\mathcal{V}$.

To extract 2D representations from this 3D model, we define a rendering function $\mathcal{R}$ that projects the 3D mesh onto a 2D plane given camera parameters $\phi = (R, t, K)$, where $R \in \mathbb{R}^{3 \times 3}$ is the rotation matrix, $t \in \mathbb{R}^3$ is the translation vector, and $K \in \mathbb{R}^{3 \times 3}$ is the intrinsic camera matrix:
\begin{equation}
\{I_d, I_n, I_s\} \leftarrow \mathcal{R}(\{\mathcal{V}, \mathcal{F}\}, \phi) ,
\end{equation}
where $I_d \in \mathbb{R}^{H \times W \times 1}$ is the depth map, $I_n \in \mathbb{R}^{H \times W \times 3}$ is the surface normal map, and $I_s \in \mathbb{R}^{H \times W \times 3}$ is the skeleton representation with $J$ joints.

To enrich the camera viewpoint information, we incorporate depth maps. Additionally, surface normals from SMPL are used to capture detailed human surface characteristics, enhancing the precision of the generated augmentations. As a result, depth maps, surface normals, and human skeletons from SMPL, along with a reference image to control identity, are fed into the generative model as guidance.

\subsection{Pose and Viewpoint Diversification}\label{sec:augmentation}
To mitigate the biased camera viewpoint and human pose in training dataset, we augment images with \textit{uniformly distributed camera viewpoint} and \textit{human poses sourced from outside the training dataset}. For camera viewpoints, we augment the images adjusting two factors: elevation and azimuth.

We sample the elevation angle $\alpha$ from a uniform distribution with the hyperparameters $\alpha_{\mathrm{min}}$ and $\alpha_{\mathrm{max}}$, denoted as $\alpha \sim \mathcal{U}(\alpha_{\mathrm{min}}, \alpha_{\mathrm{max}})$, assuming that the camera is not positioned below the ground and that a person becomes indistinguishable when the camera is positioned above a certain degree. Similarly, for the azimuth angle $\gamma$, we uniformly sample within the bounds $\gamma_\mathrm{{min}}$ and $ \gamma_{\mathrm{max}}$, represented as  $\gamma \sim \mathcal{U}(\gamma_{\mathrm{min}}, \gamma_{\mathrm{max}})$. In addition, the camera is always directed towards the center of the human mesh. This approach allows for the capture of a person from any direction. With these considerations, we found that the distribution of camera viewpoint is significantly unbiased, as depicted in Figure~\ref{fig:multi-cost}.

For human poses, we source them from outside the training dataset for diversification. As Re-ID datasets often contain similar human poses, solely training on these datasets can lead to overfitting and limited generalization capabilities. Let $\mathcal{P}_{\mathrm{ext}}$ represent the set of human poses extracted from external sources (e.g., dance videos~\cite{chan2019everybody}). For each augmentation, we randomly sample a pose $\theta \sim \mathcal{P}_{\mathrm{ext}}$ from this external distribution uniformly. By incorporating a wide range of external poses, our approach improves the model's ability to handle unseen poses that are not present in the training dataset.
Using these diversified viewpoint and pose distributions, we render SMPL models into 2D representations according to:
\begin{equation}
\{I_d, I_n, I_s\} = \mathcal{R}(\mathcal{M}(\beta, \theta), \phi),
\end{equation}
where $\phi = (R(\alpha, \gamma), t, K)$, $R(\alpha, \gamma)$ represents rotation matrix derived from azimuth and elevation, $\alpha \sim \mathcal{U}(\alpha_{min}, \alpha_{max})$, and $\gamma \sim \mathcal{U}(\gamma_{min}, \gamma_{max})$, respectively. These rendered 2D representations act as input conditions for the generative model, providing guidance for the data generation process. With the augmented training dataset from generative model, we can train arbitrary Re-ID models with robustness to camera viewpoint and human pose variations. 

\subsection{Pose-Diversified Augmentation}\label{sec:diffusion}
We generate training images with diverse camera viewpoints and human poses to reduce bias in the distribution of the training data. However, when training the generative model on a human Re-ID dataset without careful consideration, it may produce poor results for camera viewpoints or human poses that are rarely present in the training dataset, as the quality of the generated data is limited by the capabilities of the generative model. If the generative model is unable to handle out-of-distribution human poses not seen during training, it will produce degraded training data. 

In this work, we address this problem by leveraging the extensive knowledge in pre-trained Stable Diffusion (SD)~\cite{rombach2022high}. Specifically, we fine-tune SD to accommodate rendered pose conditions, adapting the framework proposed by \citet{AnimateAnyone}. This approach effectively preserves the identity of the input image while taking advantage of Stable Diffusion's comprehensive pre-trained knowledge.

Let $x_0$ be the target image we want to generate, and $x_T$ be pure Gaussian noise. The forward process gradually adds noise to the image according to:
\begin{equation}
q(x_t|x_{t-1}) = \mathcal{N}(x_t; \sqrt{1-\beta_t}x_{t-1}, \beta_t\mathbf{I}),
\end{equation}
where $\beta_t$ is the noise schedule at timestep $t$.

The approach clones the pre-trained diffusion model into two branches of U-Nets. One branch, a reference U-Net $\epsilon_\theta^{ref}$, receives an image of a person whose identity is to be generated, while the other is a denoising U-Net $\epsilon_\theta$ that gradually removes Gaussian noise according to:
\begin{equation}
p_\theta(x_{t-1}|x_t, c_{\text{pose}}) = \mathcal{N}(x_{t-1}; \mu_\theta(x_t, t, c_{\text{pose}}), \Sigma_\theta(x_t, t)),
\end{equation}
where $c_{\text{pose}}$ represents the conditioning information, and $\mu_\theta$ is derived from the predicted noise $\epsilon_\theta$.

A reference U-Net provides the identity information to the denoising U-Net through an attention mechanism. Identity information is shared with the denoising U-Net within self-attention~\cite{vaswani2017attention} in each block, allowing the two parallel branches to benefit from the comprehensive pre-trained knowledge of Stable Diffusion. Formally, for each attention layer in the network:
\begin{equation}
\text{Attention}(Q, K, V) = \text{softmax}\left(\frac{QK^T}{\sqrt{d_k}}\right)V,
\end{equation}
where \(d_k\) denotes the scaling factor, the query $Q$ comes from the denoising branch, while the key $K$ and value $V$ are derived from both the reference and denoising branch to provide identity guidance while denoising.

In addition, we concatenate the depth, surface normals, and skeleton, and feed them into the pose guider network $G$, which consists of stacks of convolutional layers:
\begin{equation}
c_{\text{pose}} = G([I_d, I_n, I_s]),
\end{equation}
where \([\cdot]\) denotes concatenation. The encoded condition is then added to the projected input of the denoising diffusion model, following~\cite{AnimateAnyone}. 

The diffusion model can generate an image that retains the identity from the reference image while allowing control over its viewpoint and human pose. Augmenting the training dataset with the diffusion model conditioned on the pose distribution outlined in Sec.~\ref{sec:augmentation} helps reduce bias when training a Re-ID model. For a detailed description of the architecture, please refer to the supplementary materials.

\begin{table*}
\centering
\resizebox{0.8\linewidth}{!}{
\begin{tabular}{l|cc|cc|cc|cc}
\toprule
\multirow{2}{*}{Methods} & \multicolumn{2}{c|}{MSMT17} & \multicolumn{2}{c|}{Market1501} & \multicolumn{2}{c|}{CUHK03 (D)} & \multicolumn{2}{c}{CUHK03 (L)} \\
 & \multicolumn{1}{c}{mAP $\uparrow$} & \multicolumn{1}{c|}{R1 $\uparrow$} & \multicolumn{1}{c}{mAP $\uparrow$} & \multicolumn{1}{c|}{R1 $\uparrow$} & \multicolumn{1}{c}{mAP $\uparrow$} & \multicolumn{1}{c|}{R1 $\uparrow$} & \multicolumn{1}{c}{mAP $\uparrow$} & \multicolumn{1}{c}{R1 $\uparrow$} \\ \midrule\midrule

  TransReID~\cite{he2021transreid} & \underline{69.4} & 86.2 & 89.5 & 95.2 & - & - & - & - \\
  AAFormer~\cite{zhu2023aaformer}  & 65.6 & 84.4 & 88.0 & 95.4 & 77.2 & 78.1 & 79.0 & 80.3 \\ 
  AGW~\cite{ye2021deep} & 49.3 & 68.3 & 87.8 & 95.1 & - & - & 62.0 & 63.6 \\ 
  FlipReID~\cite{ni2021flipreid} & 68.0 & 85.6 & 89.6 & 95.5 & - & - & - & - \\
  CAL~\cite{rao2021cal} & 64.0 & 84.2 & 89.5 & 95.5 & - & - & - & - \\
  PFD~\cite{wang2022pose} & 64.4 & 83.8 & 89.7 & 95.5 & - & - & - & - \\
  SAN~\cite{jin2020san} & 55.7 & 79.2 & 88.0 & 96.1 & 74.6 & 79.4 & 76.4 & 80.1 \\
  LDS~\cite{zang2021learning} & 67.2 & \underline{86.5} & 90.4 & 95.8 & - & - & - & - \\

  MPN~\cite{ding2020multi} & 62.7 & 83.5 & 90.1 & \underline{96.4} & 79.1 & 83.4 & 81.1 & 85.0 \\
  MSINet~\cite{gu2023msinet} & 59.6 & 81.0 & 89.6 & 95.3 & - & - & - & - \\
  SCSN~\cite{chen2020salience} & 58.5 & 83.8 & 88.5 & 95.7 & 81.0 & 84.7 & 84.0 & 86.8 \\
  \midrule

Baseline (CLIP-reID~\cite{CLIP-reID}) & 68.0 & 85.8 & 89.6 & 95.5 & 93.7 & 95.5 & 95.5 & 96.6 \\
  \hlrow\textbf{+ \ours } & \textbf{71.0} & \textbf{87.5} & 90.3 & 95.6 & 95.5 & \textbf{97.4} & 97.2 & \underline{97.8} \\ \midrule

  Baseline (SOLIDER~\cite{SOLIDER}) & 67.4 & 85.9 & \underline{91.6} & 96.1 & \underline{95.6} & 96.7 & \underline{97.4} & \textbf{98.5} \\
  \hlrow\textbf{+ \ours } & 68.3 & 85.9 & \textbf{92.3} & \textbf{96.6} & \textbf{96.2} & \underline{97.2} & \textbf{97.6} & \textbf{98.5} \\

\bottomrule
\end{tabular}
}
\vspace{-5pt}
\caption{\textbf{Quantitative comparison on standard Re-ID benchmarks.} Since our generative augmentation can be applied to any Re-ID model, we trained two recent state-of-the-art baselines~\cite{CLIP-reID,SOLIDER} with \ours.}
\label{tab:baseline_comp}

\vspace{-10pt}
\end{table*}

\section{Experiments}

\subsection{Implementation Details}
We use CLIP-reID~\cite{CLIP-reID} and SOLIDER~\cite{SOLIDER} as the baselines to validate the effectiveness of the \ours augmented dataset. The entire training process is divided into three parts: training the generative model, generating images for augmentation, and training the baseline Re-ID models using the augmented dataset.

\paragrapht{Step 1: Training of generative model.} The training of our generative model involves two stages. First, it learns a general human representation using a fashion video dataset~\cite{zablotskaia2019dwnet}, followed by fine-tuning on person Re-ID datasets. Throughout both stages, the weights of the autoencoders~\cite{yu2021vector} and the CLIP~\cite{radford2021learning} image encoder are kept frozen, focusing on the learning of reference U-Net, denoising U-Net, and pose guider. Initially, the reference and denoising U-Nets are initialized from the Stable Diffusion~\cite{rombach2022high} model and fine-tuned with the fashion video dataset. We employ Mean Squared Error (MSE) loss, optimized using Adam~\cite{kingma2014adam} with a learning rate of 1e-5 and weight decay of 0.01. The first stage takes approximately 15 hours on a single NVIDIA RTX A6000 GPU with a batch size of 2. In the second stage, the model is further fine-tuned using Re-ID datasets consisting of pedestrian images from CCTV cameras, with the images resized to 192\(\times\)384 and the batch size increased to 4.

\paragrapht{Step 2: Augmentation with generative model.} Our approach leverages data augmentation via a generative model to enrich the training data for the Re-ID model. We begin by rendering SMPLs extracted from the Everybody Dance Now dataset~\cite{chan2019everybody}, an external dataset not used during Re-ID model training. From these SMPLs, we generate skeleton, depth, and normal maps using various camera viewpoints and a wide range of human poses, creating a ``target condition gallery."
For each person identity (PID), we randomly select one instance and pair it with a randomly chosen target condition from this gallery. This pair is then fed into the generative model to produce a new image. We repeat this process iteratively, ensuring the same number of iterations for each PID. These generated images are then combined with the original training datasets for training.
The camera viewpoints of generated images are controlled by the azimuth $\gamma$ and elevation angles $\alpha$, where the angles are sampled within a limited range $\alpha_{min} = 0^\circ, \alpha_{max} = 30^\circ, \gamma_{min} = 0^\circ,$ and $ \gamma_{max} = 360^\circ$.

\paragrapht{Step 3: Training baseline Re-ID models.}
We trained two baseline Re-ID models, CLIP-reID and SOLIDER, using both the original and augmented Re-ID datasets. For training the baselines, we followed the procedure outlined in their respective papers. 

\subsection{Evaluation Protocol}
We evaluate the performance of the models using four publicly available person Re-ID datasets, \textit{i.e.}, MSMT17~\cite{MSMT17}, Market-1501~\cite{Market1501}, CUHK03~(L), and CUHK03~(D)~\cite{CUHK03-1}. L and D stand for Labeled and Detected, respectively. For evaluation metrics, we use two standard Re-ID metrics: cumulative matching characteristics at Rank-1~(R1) and mean average precision~(mAP). 

\begin{figure}[t]
  \centering
\includegraphics[width=1.0\linewidth]{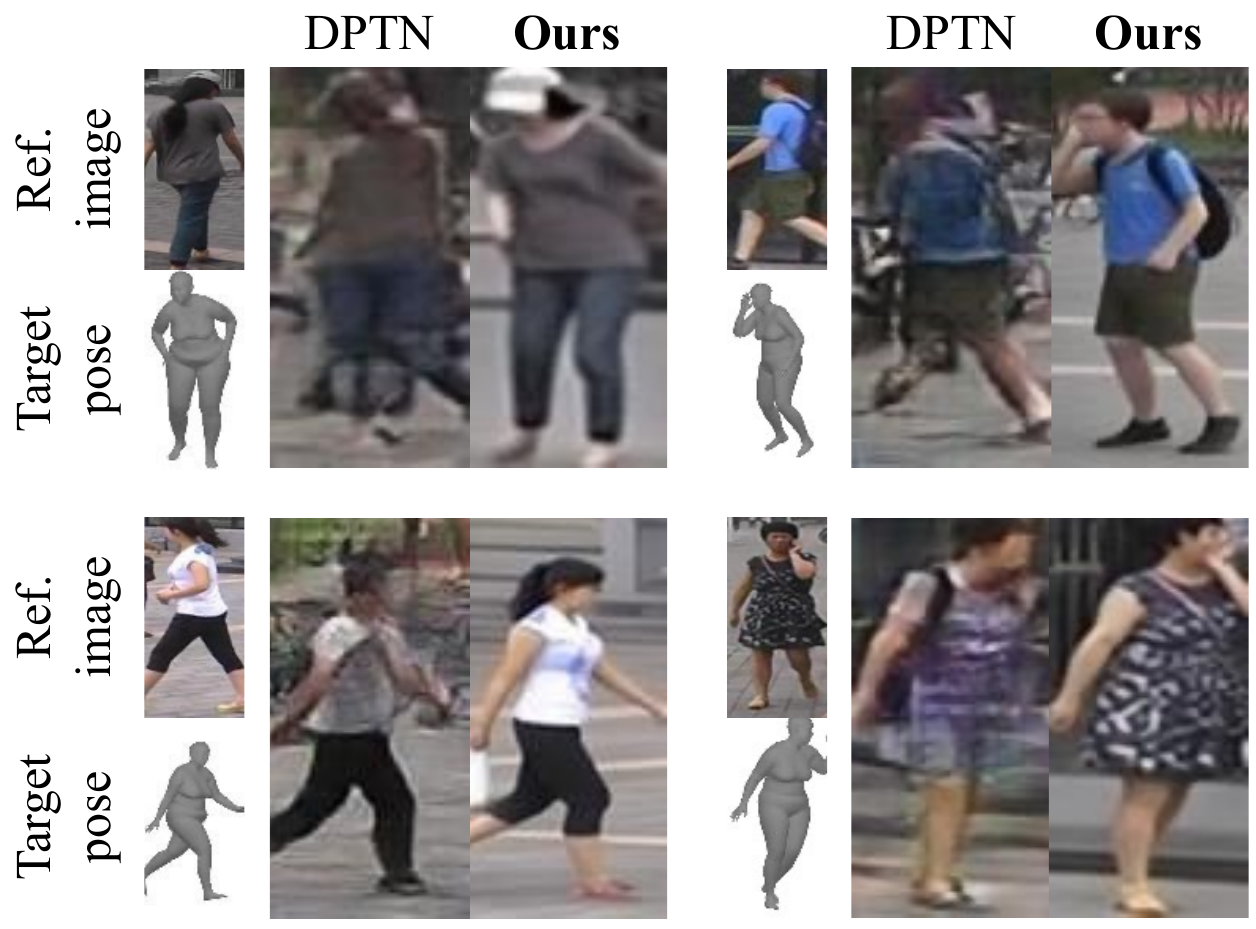}
\\
\vspace{-5pt}
\caption{\textbf{Qualitative comparison.} We compare our generated output with DPTN~\cite{zhang2022exploring}, showing that \ours can generate more realistic images while better preserving identity (e.g., cap and backpack) and accurately following the target pose.}
\label{fig:qualitative-comparison}
\vspace{-15pt}
\end{figure}

\begin{table*}[t]
\centering
\resizebox{0.8\textwidth}{!}{
\begin{tabular}{c|c|c|cc|cc|cc|cc}
\toprule
& Human Pose & Viewpoint & \multicolumn{2}{c|}{MSMT17} & \multicolumn{2}{c|}{Market-1501} & \multicolumn{2}{c|}{CUHK03~(D)} & \multicolumn{2}{c}{CUHK03~(L)} \\
& Augmentation & Augmentation & mAP & R1 & mAP & R1 & mAP & R1 & mAP & R1 \\
\midrule\midrule
\textbf{(I)} & \xmark & \xmark & 68.0 & 85.8 & 89.6 & 95.5 & 93.7 & 95.5 & 95.5 & 96.6 \\
\textbf{(II)} & \xmark & \cmark & \underline{70.9} & 87.0 & 90.1 & 95.3 & 93.8 & 95.3 & 95.8 & 96.8 \\
\textbf{(III)} & \cmark & \xmark & \underline{70.9} & \underline{87.3} & \underline{90.2} & \underline{95.4} & \underline{94.6} & \underline{96.6} & \underline{96.4} & \underline{97.4} \\
\hlrow\textbf{(IV)} & \cmark & \cmark & \textbf{71.0} & \textbf{87.5} & \textbf{90.3} & \textbf{95.6} & \textbf{95.5} & \textbf{97.4} & \textbf{97.2} & \textbf{97.8} \\
\bottomrule
\end{tabular}
}
\vspace{-5pt}
\caption{\textbf{Quantitative validation of \ours augmentation strategies.} We progressively apply human pose and viewpoint augmentation starting from the baseline. The results demonstrate that both types of augmentation independently enhance performance and provide an even greater improvement when combined.}
\label{tab:ablation}
\vspace{-15pt}
\end{table*}

\begin{table}[t]
\centering
\resizebox{\linewidth}{!}{
\begin{tabular}{c|l|c|c|cc|cc}
\toprule
&\multirow{3}{*}{Training Dataset}  & \multirow{3}{*}{\# of Images} & \multirow{3}{*}{PIDs} & \multicolumn{4}{c}{Market1501} \\
&&&&\multicolumn{2}{c|}{ResNet-50}&\multicolumn{2}{c}{SOLIDER}\\
&&&& mAP $\uparrow$ & R1 $\uparrow$ & mAP $\uparrow$ & R1 $\uparrow$ \\ \midrule\midrule
 \textbf{(I)}&Baseline Dataset & 11,883 & 619 & 74.7 & 88.9 & 91.6 & 96.1 \\\midrule
 \textbf{(II)}& \textbf{(I)} + Real Images & 30,453 & 619 &  \underline{77.8} & \underline{92.8} & \underline{91.8} & \underline{96.4} \\
 \hlrow\textbf{(III)}& \textbf{(I) + \ours Augmented} & 30,453 & 619 &\textbf{80.2} & \textbf{92.9} &\textbf{92.3} & \textbf{96.6}\\ 

\bottomrule
\end{tabular}
}
\vspace{-5pt}
\caption{\textbf{Ablation on pose and viewpoint diversity with fixed data size.} Our augmentation strategy achieves better performance gains than simply increasing the dataset with additional in-domain real images.}
\label{tab:abl_mars}

\vspace{-10pt}
\end{table}

\begin{table}[t]
\centering
\resizebox{\linewidth}{!}{
\begin{tabular}{c|l|c|c|cc|cc}
\toprule
&\multirow{3}{*}{Training Dataset}  & \multirow{3}{*}{\# of Images} & \multirow{3}{*}{PIDs} & \multicolumn{4}{c}{Market1501} \\
&&&&\multicolumn{2}{c|}{ResNet-50}&\multicolumn{2}{c}{SOLIDER}\\
&&&& mAP $\uparrow$ & R1 $\uparrow$ & mAP $\uparrow$ & R1 $\uparrow$ \\ \midrule\midrule
 \textbf{(I)}&Baseline Dataset & 11,883 & 619 & 74.7 & 88.9 & 91.6 & 96.1 \\\midrule
 \textbf{(II)}&MARS & 495,857 & 619 & 72.4 & 82.3 &  88.2 & 89.3 \\
 \hlrow\textbf{(III)}& \textbf{(I) + \ours Augmented} & 30,453 & 619 &\textbf{80.2} & \textbf{92.9} & \textbf{92.3} & \textbf{96.6} \\ 

\bottomrule
\end{tabular}
}
\vspace{-5pt}
\caption{\textbf{Comparison with large-scale in-domain real image augmented dataset.} Increasing the dataset size alone, without ensuring diversity in human poses, does not guarantee improved performance.}
\label{tab:abl_mars2}

\vspace{-20pt}
\end{table}

\subsection{Quantitative Comparisons}

The quantitative results, presented in Table~\ref{tab:baseline_comp}, demonstrate the effectiveness of our approach. To emphasize its applicability to arbitrary Re-ID models, we conducted experiments on two state-of-the-art models~\cite{CLIP-reID, SOLIDER}. Both models exhibited significant performance boosts when trained on a dataset augmented with our approach. This validates the broad applicability of our augmentation method to a variety of Re-ID models. Furthermore, we observed consistent performance improvements across various datasets, further validating our augmentation framework and highlighting the efficacy of addressing sparsely distributed human poses and viewpoints.

\subsection{Qualitative Results}

\paragraph{Qualitative comparisons.} In Figure~\ref{fig:qualitative-comparison}, we present a qualitative comparison to a recent GAN-based approach~\cite{zhang2022exploring}. In contrast to GAN-based methods, our approach effectively generates poses sourced from outside the dataset. GAN-based methods struggle to generalize to diverse poses, often resulting in blurry outputs and a limited ability to maintain the identity of both the reference and the target pose, particularly when dealing with complex features such as accessories, specific clothing details, or bags. This underscores the advantage of utilizing a pre-trained diffusion model, which possesses significant general knowledge about the world.

\begin{figure}[t]
  \centering
\includegraphics[width=1.0\linewidth]{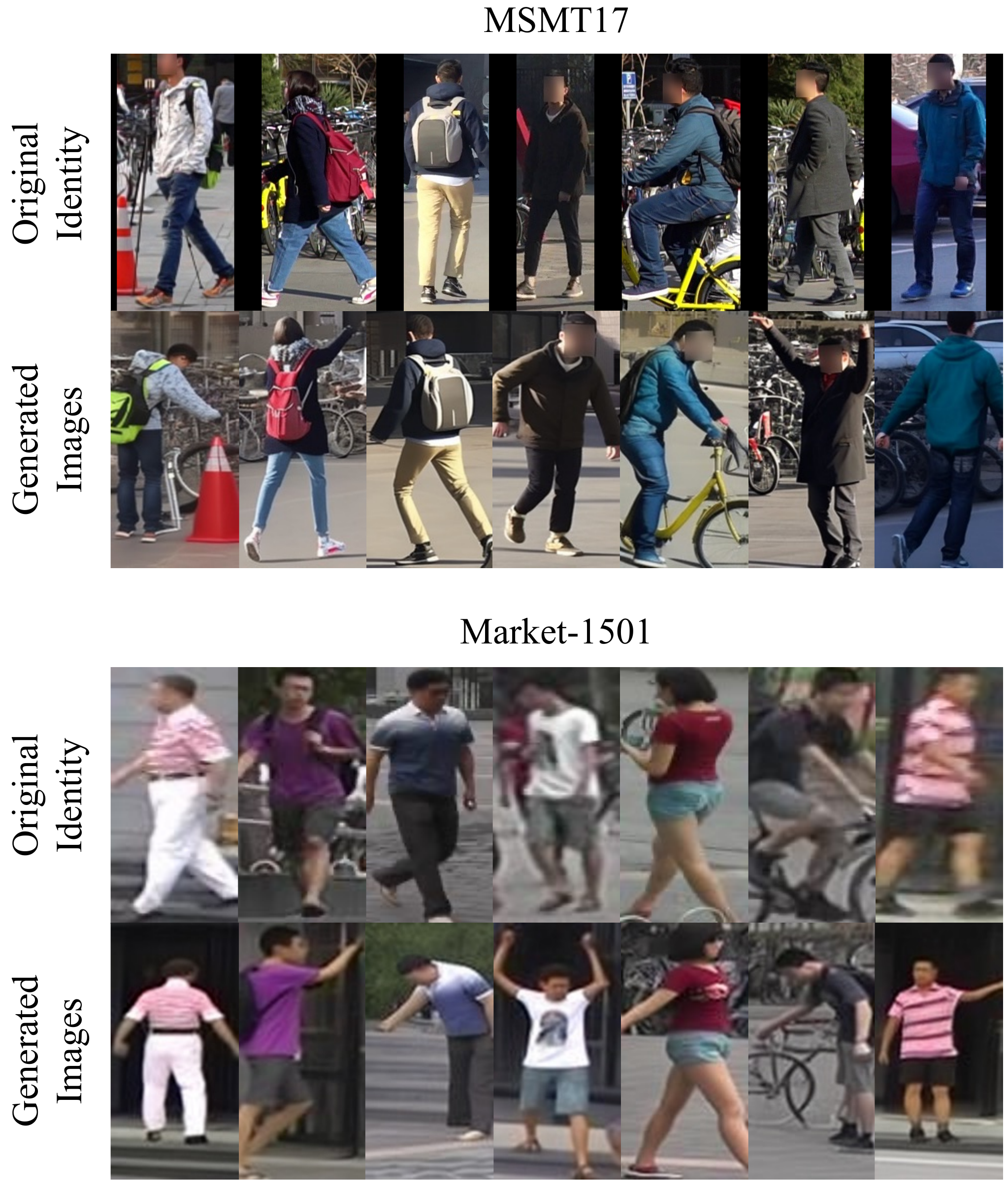}
\\
\vspace{-5pt}
\caption{\textbf{Qualitative results.} Example images from the augmented MSMT17 and Market-1501 dataset demonstrate how the generated images preserve original identities while maintaining realism and consistency with the Re-ID dataset.}
\label{fig:visualization}
\vspace{-15pt}
\end{figure}

\paragrapht{Visualization of generated data.} In Figure~\ref{fig:visualization}, we present visualization results on two datasets, MSMT17~\cite{MSMT17} and Market-1501~\cite{Market1501}. Given a reference image, our method faithfully preserves its identity while being able to generate diverse poses with high fidelity.

\subsection{Ablation Study and Analysis} 

\paragraph{Ablation on the \ours augmentation strategy.}
In Table~\ref{tab:ablation}, we conduct an ablation study using CLIP-ReID~\cite{CLIP-reID} to verify the effectiveness of our viewpoint and human pose augmentation. \textbf{(I)} serves as the baseline, representing a Re-ID model trained on the original dataset without our augmentation. For \textbf{(II)} and \textbf{(III)}, we augment viewpoints and human poses, respectively. Both augmentations demonstrate significant performance gains, validating the effectiveness of targeting the biased distributions of human pose and viewpoint for augmentation. \textbf{(IV)} demonstrates the full augmentation strategy of our model. The performance gains observed in \textbf{(IV)} compared to both \textbf{(II)} and \textbf{(III)} confirm that both types of augmentation are not only beneficial individually, but also exhibit a complementary effect when combined, leading to further improvements in performance.

\paragrapht{Comparison with real image augmented dataset.}
In this analysis, we investigated whether the performance gain from augmentation originates from an increased dataset size or from increased pose diversity. To validate this, we compared our augmentation method with real-world data collection as shown in Table~\ref{tab:abl_mars} and Table~\ref{tab:abl_mars2}. MARS~\cite{zheng2016mars} is an extended version of Market-1501, which samples more images from the same videos used in Market-1501. For simplicity, we conduct experiments using ResNet-50.

In Table~\ref{tab:abl_mars}, the baseline training dataset, \textbf{(I)}, is a subset of Market-1501, filtering out identities not present in MARS. \textbf{(II)} is the subset of the MARS dataset, matching the number of training images with ours.
\textbf{(III)} is the dataset augmented with our approach. The performance gap between \textbf{(II)} and \textbf{(III)}, despite both being trained on datasets of the same size, demonstrates that merely increasing the dataset without introducing diversity in pose and viewpoint results in suboptimal performance. 

\begin{figure}
\centering
\includegraphics[width=0.8\linewidth]{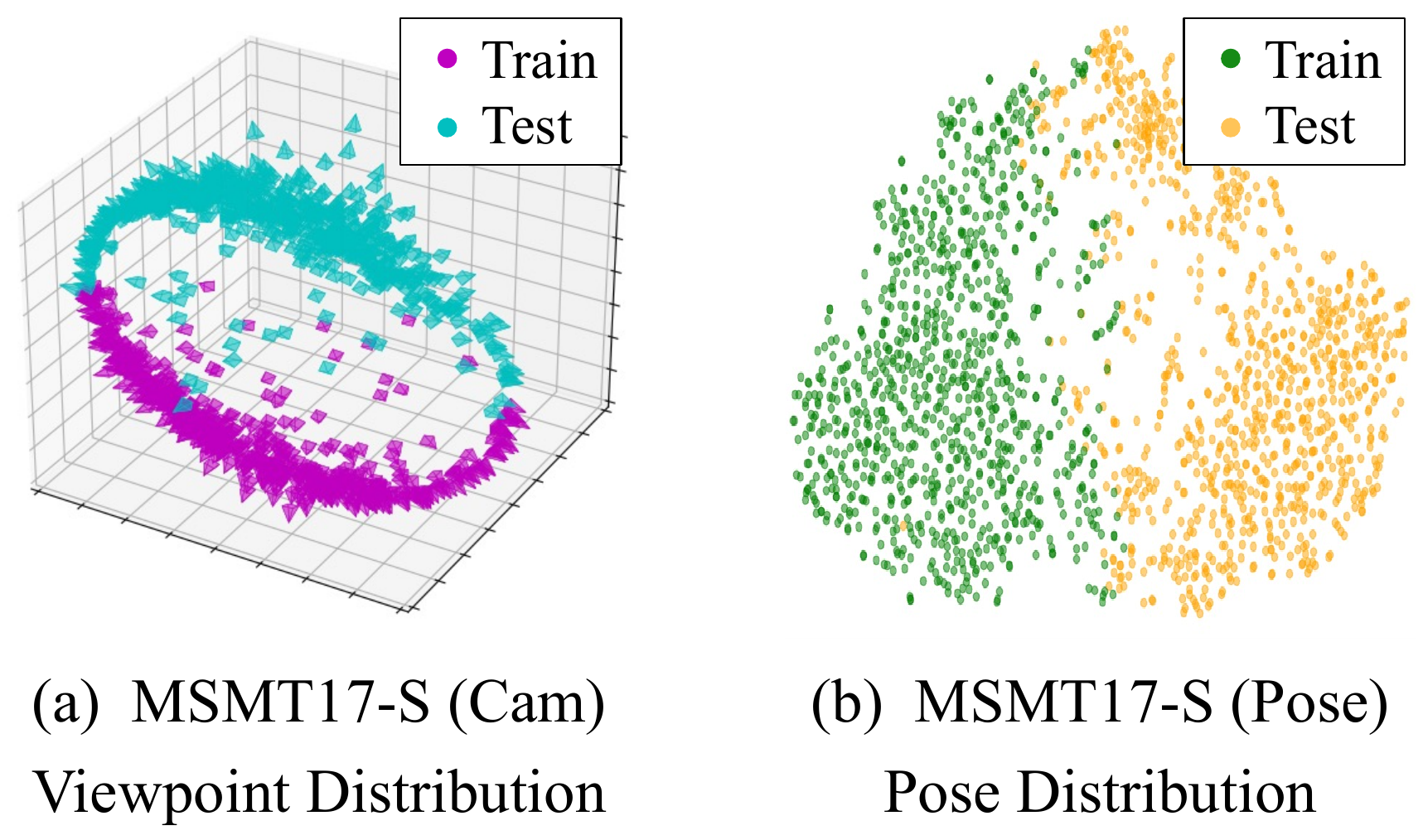}
\caption{\textbf{Visualization of the split data.} To validate the generalization power of our framework, we split the MSMT17 dataset into train/test sets using two distinct approaches: (a) splitting based on viewpoint, and (b) splitting based on human pose. The visualization clearly illustrates the separation between the train and test distributions.}

\label{fig:split_msmt}
\vspace{-10pt}
\end{figure}
\begin{table}[t]

\centering

\resizebox{\linewidth}{!}{
\begin{tabular}{l|cc|cc}  
\toprule
\multirow{2}{*}{Method} & \multicolumn{2}{c|}{MSMT17-S (Cam)} & \multicolumn{2}{c}{MSMT17-S (Pose)} \\
 & \multicolumn{1}{c}{mAP $\uparrow$} & \multicolumn{1}{c|}{R1 $\uparrow$} & \multicolumn{1}{c}{mAP $\uparrow$} & \multicolumn{1}{c}{R1 $\uparrow$} \\ \midrule\midrule

 Baseline (SOLIDER~\cite{SOLIDER}) & \underline{46.6} & \underline{66.7} & \underline{60.0} & \underline{76.0} \\
 \hlrow\textbf{+ \ours} & \textbf{52.9} & \textbf{71.1} & \textbf{63.7} & \textbf{78.4} \\

\bottomrule
\end{tabular}
}
\vspace{-5pt}
\caption{\textbf{Generalization performance on custom split.} \textit{Cam} denotes dataset split with camera viewpoint, while \textit{Pose} denotes dataset split with human pose. \ours demonstrates generalization even on extreme camera viewpoints and human poses not present in the training set. 
}
\label{tab:split_data}

\vspace{-15pt}
\end{table}

In Table~\ref{tab:abl_mars2}, although the real-image augmented dataset \textbf{(II)} is approximately 16 times larger than our augmented dataset \textbf{(III)}, the Re-ID model trained on \textbf{(II)} performs worse on both metrics compared to the model trained on \textbf{(III)}. This highlights that Re-ID model performance is not solely dependent on dataset size but is significantly influenced by the diversity and generalization capability of the images within the dataset.

\paragrapht{Analysis on generalization effectiveness.}

To rigorously assess the generalization capabilities of our framework, we devised an experiment involving an extreme dataset split of the MSMT17 dataset, intentionally amplifying the bias of human pose and viewpoint in the training dataset. We propose two distinct split variants: MSMT17-S (Cam) and MSMT17-S (Pose), where \textit{Cam} divides the dataset into train and test sets with non-overlapping camera viewpoints, while \textit{Pose} divides it based on human pose. These datasets create clear experimental conditions by intentionally partitioning the dataset according to viewpoints and human poses, as visualized in Figure~\ref{fig:split_msmt}. This figure provides a clear visual representation of how the data was split and the lack of overlap between the training and test sets for both viewpoints and human poses. For each split, we fine-tune the diffusion model on the training split and generate the dataset. Then, we use the generated dataset to train the baseline model. We select SOLIDER as the baseline model for this experiment. 

As shown in Table~\ref{tab:split_data}, the results clearly demonstrate that our approach significantly improves generalization. The performance enhancements on these filtered datasets were substantially higher, indicating the effectiveness of our method in diversifying the viewpoints and human poses of the original dataset, thereby improving the model's ability to generalize to new, unseen poses, even in this extreme setup. For further details on the dataset split process, please refer to the supplementary material.

\paragrapht{Analysis on performance improvements on curated non-pedestrian dataset.}
We further validate our approach by creating a test dataset that includes a video exhibiting abnormal behaviors, as detailed in Table~\ref{tab:non_ped_evaluation}. Unlike traditional pedestrian-centric re-identification datasets, the curated test dataset comprises dynamic images captured in an indoor setting using eight cameras, with an emphasis on abnormal behaviors such as violence and self-harm, which introduce significant diversity in human pose. This dataset includes 23,399 instances across 127 person IDs (PIDs). For our baseline, we employed a ResNet50 model pretrained on LUPerson~\cite{LU-unsuper}.

As shown in Table~\ref{tab:non_ped_evaluation}, our method improved performance on both the existing benchmark dataset Market-1501 and abnormal behaviors CCTV datasets. However, the improvement was significantly larger on our curated test dataset ($+$13.6 mAP) compared to Market-1501 ($+$5.6 mAP). This substantial improvement under zero-shot conditions demonstrates the robustness of our method in generalizing to new poses and camera distributions.

\begin{table}[t]
\centering
\resizebox{\linewidth}{!}{
\begin{tabular}{l|cc|cc}  
\toprule
\multirow{2}{*}{Method} & \multicolumn{2}{c|}{Market1501} & \multicolumn{2}{c}{Non-pedestrian Dataset} \\
 & mAP $\uparrow$ & R1 $\uparrow$ & mAP $\uparrow$ & R1 $\uparrow$ \\ 
\midrule\midrule
Baseline (LUPerson~\cite{LU-unsuper}) & \underline{83.0} & \underline{93.0} & \underline{75.7} & \underline{84.2} \\
\hlrow \textbf{+ \ours} & \textbf{88.6 (\textcolor{ForestGreen}{$+$5.6})} & \textbf{95.0 (\textcolor{ForestGreen}{$+$2.0})} & \textbf{89.3 (\textcolor{ForestGreen}{$+$13.6})} & \textbf{95.2 (\textcolor{ForestGreen}{$+$11.0})} \\
\bottomrule
\end{tabular}
}
\caption{\textbf{Generalization performance on non-pedestrian test dataset.} We curated a non-pedestrian test dataset consisting of videos that depict unusual human behaviors. The results further validate the robustness of \ours in handling diverse human poses.}
\label{tab:non_ped_evaluation}
\vspace{-12pt}
\end{table}
\begin{table}

\centering

\resizebox{0.6\linewidth}{!}{
\begin{tabular}{l|cc}  
\toprule
\multirow{2}{*}{Method} & \multicolumn{2}{c}{Market-1501} \\
& \multicolumn{1}{c}{mAP $\uparrow$} & \multicolumn{1}{c}{R1 $\uparrow$} \\ \midrule\midrule

 Baseline (CBN~\cite{zhong2018camera}) & \underline{77.3} & \underline{91.3} \\
 \hlrow\textbf{+ \ours} & \textbf{82.3} & \textbf{93.4} \\ \midrule

 Baseline (BPBreID~\cite{BPBreID}) & \underline{89.4} & \underline{95.7} \\
 \hlrow\textbf{+ \ours} & \textbf{89.9} & \textbf{95.8} \\ 

\bottomrule
\end{tabular}
}
\vspace{-5pt}
\caption{\textbf{\ours applied to generalization-focused methods.} Re-ID models designed to address camera bias and pose variation benefit from the \ours augmented dataset.}
\label{tab:abl_bpb}

\vspace{-15pt}
\end{table}
\begin{table}[ht]
\centering

\begin{minipage}{.48\linewidth}
\centering
\resizebox{\linewidth}{!}{
\begin{tabular}{l|cc}
\toprule
\multirow{2}{*}{Method} & \multicolumn{2}{c}{Market-1501} \\
& mAP $\uparrow$ & R1 $\uparrow$ \\ \midrule\midrule
w/o Pre-trained Diffusion & \underline{82.7} & \underline{93.2} \\
\hlrow\textbf{w/ Pre-trained Diffusion} & \textbf{90.3} & \textbf{95.6} \\
\bottomrule
\end{tabular}
}

\centerline{(a)}
\end{minipage}\hfill
\begin{minipage}{.48\linewidth}
\centering
\resizebox{0.95\linewidth}{!}{
\begin{tabular}{l|cc}
\toprule
\multirow{2}{*}{Diffusion model trained on} & \multicolumn{2}{c}{Market1501} \\
& mAP $\uparrow$ & R1 $\uparrow$ \\ \midrule\midrule
In-domain pose and viewpoint & \underline{89.6} & \underline{95.2} \\
\hlrow\textbf{External pose and viewpoint} & \textbf{90.3} & \textbf{95.6} \\
\bottomrule
\end{tabular}
}
\centerline{(b)}
\end{minipage}
\vspace{-8pt}
\caption{\textbf{Ablation Studies.} (a) The effect of using a pre-trained diffusion model for weight initialization. (b) A comparison between diffusion models trained on in-domain versus external pose and viewpoint data. For both studies, we train the CLIP-reID model~\cite{CLIP-reID} on the generated dataset.}
\label{tab:abl_diffusion}
\vspace{-15pt}
\end{table}

\paragrapht{Analysis on the synergy of generalization-focused Re-ID models with \ours.}
In Table~\ref{tab:abl_bpb}, we explore the synergy between our approach and Re-ID models specifically designed to enhance generalization, particularly those addressing variations in camera and human poses.

We evaluate the impact of our approach on two representative models: CBN~\cite{zhong2018camera}, which addresses camera bias, and BPBreID~\cite{BPBreID}, designed to handle human pose variations. We select these models due to their public availability and proven effectiveness in their domains. While both models are already effective, they show marked improvements when trained on the dataset augmented by our approach. These results highlight that even with algorithmic advancements in pose generalization, there remains significant potential for improvement through dataset enhancements.

\paragrapht{Ablation on pre-trained diffusion model.}
In Table~\ref{tab:abl_diffusion}~(a), we conduct an ablation study to evaluate the impact of utilizing the pre-trained Stable Diffusion model~\cite{rombach2022high} for generating diverse camera viewpoints and human poses in our augmented dataset. 
Employing the pre-trained diffusion model results in a notable 7.6\% mAP improvement, underscoring the significance of foundation models in the context of dataset augmentation.

\paragrapht{Ablation on pose and viewpoint diversification.} Table~\ref{tab:abl_diffusion}~(b) presents a comparison between a model trained on generated data using in-domain poses and viewpoints and one trained using external poses and viewpoints. The model trained with pose-diversified data demonstrates a notable performance improvement, which underscores the effectiveness of our proposed method.
\section{Conclusion}
In this paper, we proposed \ours, a novel data augmentation approach that leverages pre-trained diffusion models to diversify pose and viewpoint distributions in person re-identification training datasets.
The key to success of \ours lies in diversifying human pose and viewpoint by integrating pose, viewpoint, and identity conditions into large-scale pre-trained diffusion models, effectively leveraging the vast knowledge embedded in these models to generate high-quality augmented data.
Comprehensive experiments demonstrated the effectiveness of our approach, with \ours achieving significant performance improvements compared to baselines.
\paragrapht{Acknowledgment}
This research was supported by Institute of Information \& communications Technology Planning \&
Evaluation (IITP) grant funded by the Korea government (MSIT) (RS-2019-II190075, RS-2024-00509279, RS-2025-II212068, RS-2023-00227592, RS-2025-02214479, RS-2024-00457882, RS-2025-25441838, RS-2025-25441838, RS-2025-02214479, RS-2025-02217259) and the Culture, Sports, and Tourism R\&D Program through the Korea Creative Content Agency grant funded by the Ministry of Culture, Sports and Tourism (RS-2024-00345025, RS-2024-00333068, RS-2023-00222280, RS-2023-00266509), and National Research Foundation of Korea (RS-2024-00346597).

{
    \small
    \bibliographystyle{ieeenat_fullname}
    \bibliography{main}
}

\setcounter{section}{0}
\setcounter{table}{0}
\setcounter{figure}{0}

\twocolumn[{
\begin{center}
    \Large \textbf{\ours: Pose-Diversified Augmentation for Person Re-Identification}\\[0.5em]
    \Large \textbf{-- Supplementary Materials --}\\[3em]
\end{center}
}]

\section{Detailed Explanation of MSMT17-S}
In the domain of person re-identification (Re-ID), addressing biases within datasets is crucial for improving model generalization. Such biases can lead to models that perform well under familiar conditions but struggle when faced with new scenarios. Among the various biases present, we focused on addressing camera viewpoint bias and human pose bias inherent in the dataset through dataset augmentation. To rigorously evaluate and challenge the generalization capabilities of our \ours method, we create two specialized versions of the MSMT17 dataset: MSMT17-S (Cam) and MSMT17-S (Pose).

MSMT17-S (Cam) is specifically designed to amplify the bias related to camera viewpoints. In this version of the MSMT17 dataset, the training set consists of images captured from a specific set of camera angles, while the test set includes entirely different angles that do not overlap with those in the training data. This deliberate split maximizes the viewpoint bias in the training data, creating a scenario where the model must generalize to new, unseen camera angles during testing. MSMT17-S (Cam) serves as a challenging benchmark to evaluate whether the \ours augmentation method can effectively help models overcome viewpoint biases, which are prevalent in real-world Re-ID scenarios where camera placements vary widely.

MSMT17-S (Pose), on the other hand, is designed to emphasize the bias related to human poses. In this dataset, the training data is composed of individuals captured in a specific range of poses, while the test set features entirely different poses that are not present in the training data. By maximizing pose bias in this way, MSMT17-S (Pose) creates a challenging environment where a Re-ID model must generalize to new poses that it has never seen during training. This setup is particularly relevant for testing how well \ours can assist a model in adapting to a wide variety of human postures, which is crucial for accurate person identification across different activities and positions.

These datasets specifically highlight the challenges posed by biases in camera viewpoints and human poses, respectively, allowing us to assess the effectiveness of the \ours augmentation strategy in mitigating these biases.

\section{Conditional Diffusion Models for Pose-Diversified Augmentation}
\begin{figure*}[t]
\begin{center}
\includegraphics[width=1\textwidth]{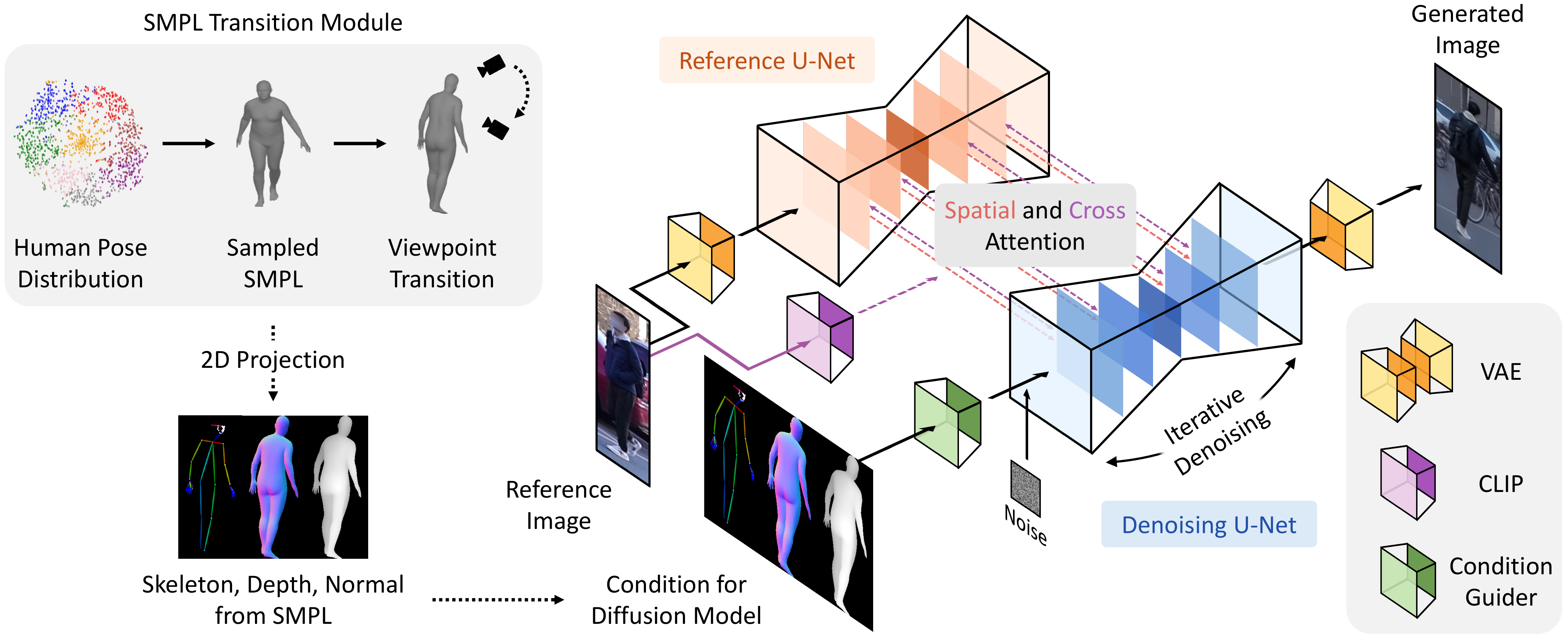}
\end{center}
\caption{\textbf{Overall architecture of generative model in \ours.}
Given the viewpoint and pose distributions, we first render the body shape sampled from the distribution using SMPL, generating the corresponding skeleton, depth map, and normal maps. These conditions, along with a reference image for identity preservation, are then fed into generative module, which consists of two branches: the reference U-Net processes the identity information from the reference image, while the denoising U-Net generates a person with the same identity, given the input conditions. The denoising U-Net generates images by iterating through the denoising process.}
\label{supl:diffusion_model}
\vspace{-10pt}
\end{figure*}

In this section, we delineate the architecture of the conditional diffusion model we employed. This model is designed to generate images with diverse camera viewpoints and human poses, conditioned on the relevant input data. However, naively training the generative model on a human Re-ID dataset without careful consideration may produce degenerated results for camera viewpoints or human poses that are rarely present in the training dataset. 

We address this problem by leveraging the vast knowledge in pre-trained Stable Diffusion~\cite{rombach2022high}. We first provide a preliminary explanation of the Stable Diffusion model~\cite{rombach2022high}, followed by the method for augmenting the images in the training dataset while controlling their human pose, camera viewpoint, and identity. The overall architecture can be seen in Figure~\ref{supl:diffusion_model}.

\paragrapht{Preliminary: Stable Diffusion.} Diffusion model~\cite{ho2020denoising, rombach2022high, song2020denoising} is a generative model that samples images from the learned data distribution \(p(x)\) through iterative denoising process from Gaussian noise. Our method builds upon Stable Diffusion (SD)~\cite{rombach2022high}. SD performs a denoising process in a latent space of Autoencoder~\cite{yu2021vector}, reducing the computational cost compared to denoising in the pixel space~\cite{ho2020denoising, song2020denoising}. Specifically, the encoder in SD maps a given image \(\textbf{x}\) into a latent representation \(\textbf{z}\), denoted as \(\textbf{z} = \mathcal{E}(\textbf{x})\).

During training, SD learns a denoising U-Net~\cite{ronneberger2015u} \(\epsilon_\theta\) that predicts normally distributed noise \(\epsilon\) given a noised latent \(\textbf{z}_t\), which is a noisy latent of $\textbf{z}$ with a Gaussian noise at noise level \(t\). This U-Net function can be trained with a following objective:
\begin{equation}
\mathcal{L} = \mathbb{E}_{\mathbf{z}_t, c, \epsilon, t}\left( \Vert \epsilon - \epsilon_\theta\left( \mathbf{z}_t, c, t \right) \Vert^2_2 \right),
\end{equation}
where \(c\) denotes conditional information for generation. The condition, which is a text prompt encoded using the CLIP text encoder~\cite{radford2021learning}, enables controllability over the image generation process. The denoising U-Net is composed of three parts: downsampling block, bottleneck block, and upsampling block. Each block consists of a combination of 2D convolutional layers, self-attention layers, and cross-attention layers.

During inference, a sample \(\textbf{z}_T\) from a Gaussian distribution is gradually denoised using the trained denoising U-Net.  Undergoing the denoising process from \(t=T\) to \(t=0\), the model generates \(\textbf{z}_0\). This final latent representation is then passed through the decoder \(\mathcal{D}\) to produce the output image.

\paragrapht{Injecting human pose and camera viewpoint into diffusion model.} For conditions that should be spatially aligned with the generated output, we process them with a respective pose guider network and concatenate the processed conditions along the channel dimension. Specifically, the depth map \(\textbf{d}\in \mathbb{R}^{H\times W \times 1}\), surface normals \(\textbf{n} \in \mathbb{R}^{H\times W\times 3}\), and rendered human skeleton \(\textbf{s} \in \mathbb{R}^{H\times W \times 3}\) are each processed by a respective pose guider network. This network reduces the spatial size of the condition to \(1/8\) of the original size and embeds the pose information into \(\mathbb{R}^{\frac{H}{8}\times \frac{W}{8}\times C}\) embeddings, aligning it with the size of the latent representation in the diffusion model. The last layer of the pose guider network is initialized to zeros to minimize the initial degradation during the fine-tuning stage of the pre-trained SD model. The processed conditions are then concatenated along the channel dimension and added to the projected noise before it is fed into the U-Net.

\paragrapht{Injecting identity into diffusion model as a condition.}
For human identity, unlike the condition that is spatially aligned with the output, it is not necessarily aligned with the output. In this regard, instead of adding the condition pixel-wise, we provide the identity information to the denoising U-Net with attention. Specifically, we design a reference U-Net that has the same architecture as the denoising U-Net, while the weights of the U-Net are initialized with a pre-trained Stable Diffusion model. To inject the identity of an image into the denoising U-Net, we first feed the image into the reference U-Net. Then, the identity information is shared with the denoising U-Net using self-attention for each block. In more detail, given the intermediate feature map from the denoising U-Net \(f_1 \in \mathbb{R}^{(h \times w) \times c}\) and from the reference U-Net \(f_2 \in \mathbb{R}^{(h \times w) \times c}\), they are concatenated along the spatial dimension, followed by the self-attention layer. Then, the first half of the output is used as the input for the following layers in the denoising U-Net. In this way, the two parallel branches can benefit from the extensive pre-trained knowledge of Stable Diffusion. Additionally, the identical architecture of the two branches facilitates training by sharing the same feature space. For the cross-attention part in Stable Diffusion where text embeddings from CLIP are used, we instead utilize image embeddings from the CLIP image encoder. This is possible because both text and image embeddings are trained to reside in the same embedding space.

\section{Additional qualitative results}
\begin{figure}
\centering
\includegraphics[width=0.9\linewidth]{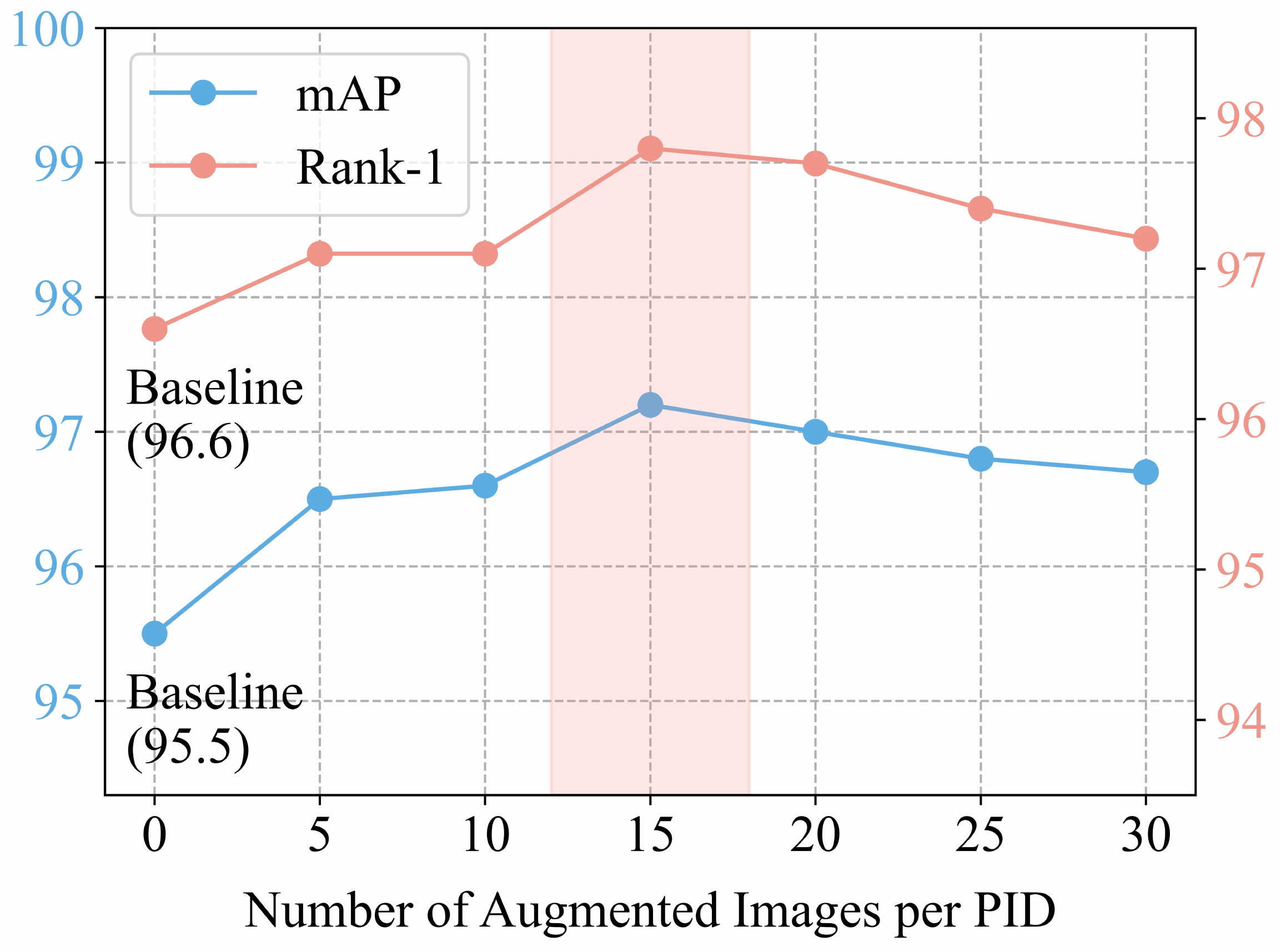}
\caption{\textbf{Impact of the number of generated images per PID.} Experiments are conducted in the \ours augmented CUHK03 (L) dataset. We use CLIP-ReID baseline.}
\label{supl:optimal_graph}
\end{figure}

\begin{figure*}
\centering
\includegraphics[width=0.85\linewidth]{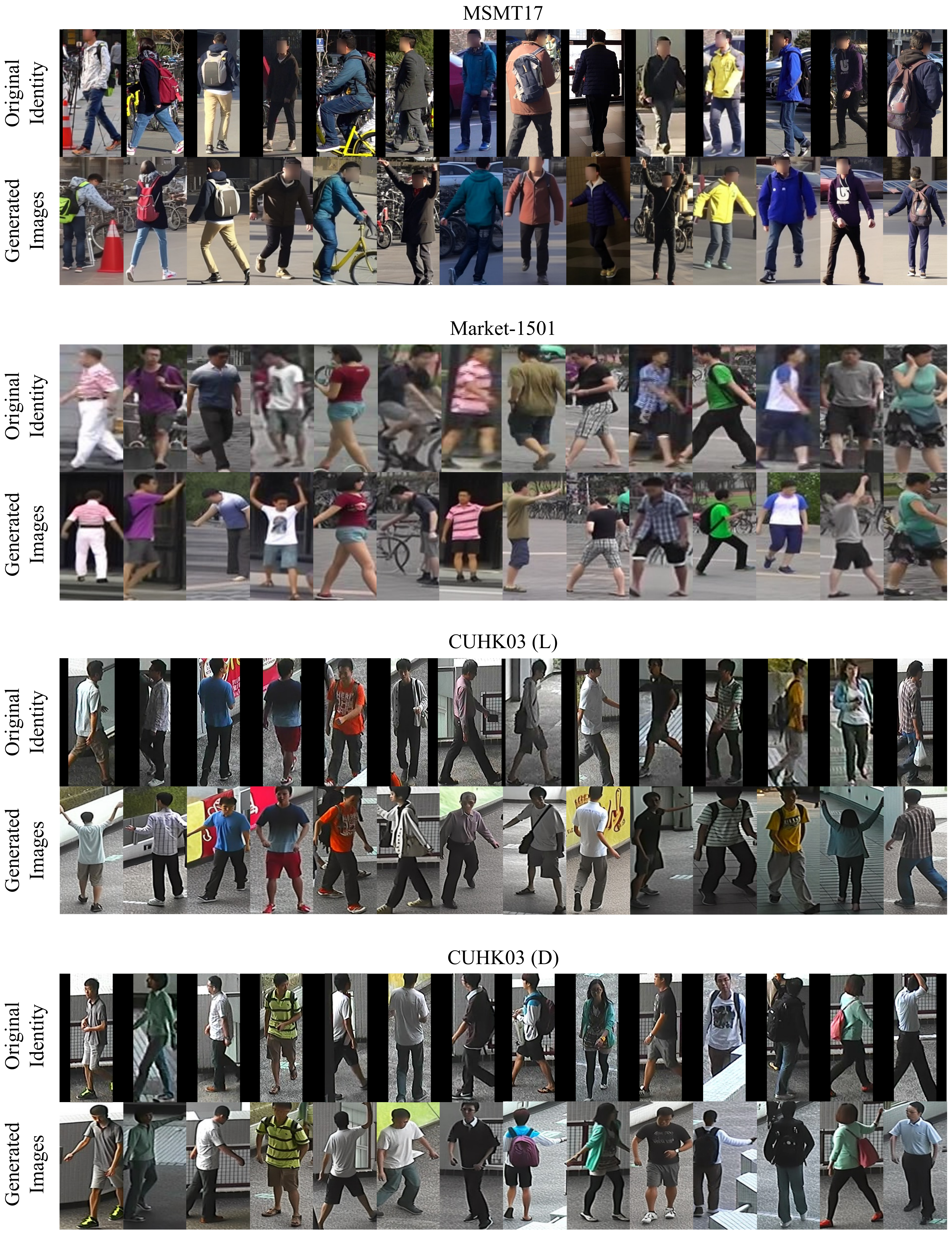}
\caption{\textbf{Additional qualitative results.} Examples of generated images from the \ours augmented datasets. The results demonstrate realistic rendering while preserving the identity of the reference images.}
\label{supl:additional_qual}
\end{figure*}
In Figure~\ref{supl:additional_qual}, we provide additional qualitative examples from the MSMT17, Market-1501, CUHK03 (D), and CUHK03 (L) datasets. The generated images maintain a high level of realism, effectively capturing the nuances of natural human appearances. Also, they successfully preserve the identity of individuals from the input reference images, ensuring that the augmented data remains faithful to the original person's characteristics. This is particularly important for re-identification tasks where identity preservation is critical. These additional results further validate the effectiveness of our proposed augmentation method across different datasets.

\section{Analysis on the number of generated images}

Figure~\ref{supl:optimal_graph} illustrates the impact of varying the number of generated images on the performance of the Re-ID model when using our proposed augmentation strategy. To assess this, we progressively increased the number of generated images in the training dataset and trained the CLIP-reID model, carefully monitoring performance changes at each increment. Note that this augmentation was applied only to the training dataset, while the test dataset remained unchanged throughout the experiments. Our findings indicate that generating approximately 15 images per person yields the highest performance for the Re-ID model.

\section{Filtering Protocols for High-Quality Augmentation}

To guarantee high-quality outputs in our augmented dataset, we applied a series of filtering procedures during both the generative model training and post-processing steps. Our filtering method uses pose scores obtained from the human pose estimation process, retaining only images that exceed certain thresholds. Filtering is applied to three types of images: \textit{reference images}, \textit{target images}, and \textit{generated images}.

Let $\mathcal{D} = \{(I_{\text{ref}}^i, I_{\text{tar}}^i)\}_{i=1}^N$ denote the data set, where $I_{\text{ref}}, I_{\text{tar}} \in \mathbb{R}^{H \times W \times C}$ represent the reference image and target image, respectively. The reference image provides the identity for the generated image, and the target image specifies the desired pose. The pose estimation model $\mathcal{P}: \mathbb{R}^{H \times W \times C} \to \mathbb{R}^K$ outputs confidence scores for $K$ keypoints:
\begin{equation}
    \mathcal{P(I)} = \{\sigma_k(I)\}_{k=1}^K,
\end{equation}
where $\sigma_k(I) \in [0, 1]$ is the confidence score for keypoint $k \in \mathcal{K}$ corresponding to each body joint.

\paragrapht{Filtering Protocol.} We use predefined thresholds $\tau$ to filter out the images that do not meet our quality criteria. First, we remove input images with heavy occlusions. For instance, when the lower body is significantly occluded, generating a full-body image from such an input becomes an ill-posed problem. Formally, we retain $I_{\text{ref}}$ if:

\begin{equation}
    \min(\{\sigma_k(I_{\text{ref}})\}_{k \in \mathcal{K}}) \geq \tau_{\text{ref}},
\end{equation}

where $\tau_{\text{ref}}$ is the threshold for the reference image. Next, we also filter out target pose images with notable occlusions, prioritizing training samples where the entire body is clearly visible. This prevents cases where the input image is clear, but the generated output unexpectedly shows occlusions. We retain $I_{\text{tar}}$ if:

\begin{equation}
    \frac{1}{N_k} \sum_{k=1}^K \sigma_k(I_{\text{tar}}) \geq \tau_{\text{tar}},
\end{equation}

where $\tau_{\text{tar}}$ is the threshold for the target image. After generating augmented images, we apply a final filtering step to discard outputs that do not align with desired poses. This ensures that the generated images are of high quality and consistent with the target pose. We first compute the mean absolute difference in confidence scores between the generated and target poses:

\begin{equation}
    \mathcal{O} = \frac{1}{N_k} \sum_{k=1}^K |\sigma_{k}(I_{\text{gen}}) - \sigma_{k}(I_{\text{tar}})|.
\end{equation}

To ensure the generated image maintains a high overall keypoint confidence score, we use $I_{\text{gen}}$ only if both conditions are satisfied:

\begin{equation}
\begin{aligned}
    \mathcal{O} \leq \epsilon_{\text{gen}} \quad \text{and} \quad \frac{1}{K} \sum_{k=1}^K \sigma_k(I_{\text{gen}}) \geq \tau_{\text{gen}},
\end{aligned}
\end{equation}

where $\epsilon_{\text{gen}}$ is a small value that determines whether the poses between the generated image and the target image are sufficiently aligned.

Through these three filtering stages, we ensure a more consistent and higher-quality augmented dataset for Re-ID training. The use of confidence scores $\sigma_k$ directly from the pose estimation model allows for effective and interpretable quality control.

\section{Analysis on Filtering Threshold Selection}

\begin{figure}[t]
  \centering
\includegraphics[width=1.0\linewidth]{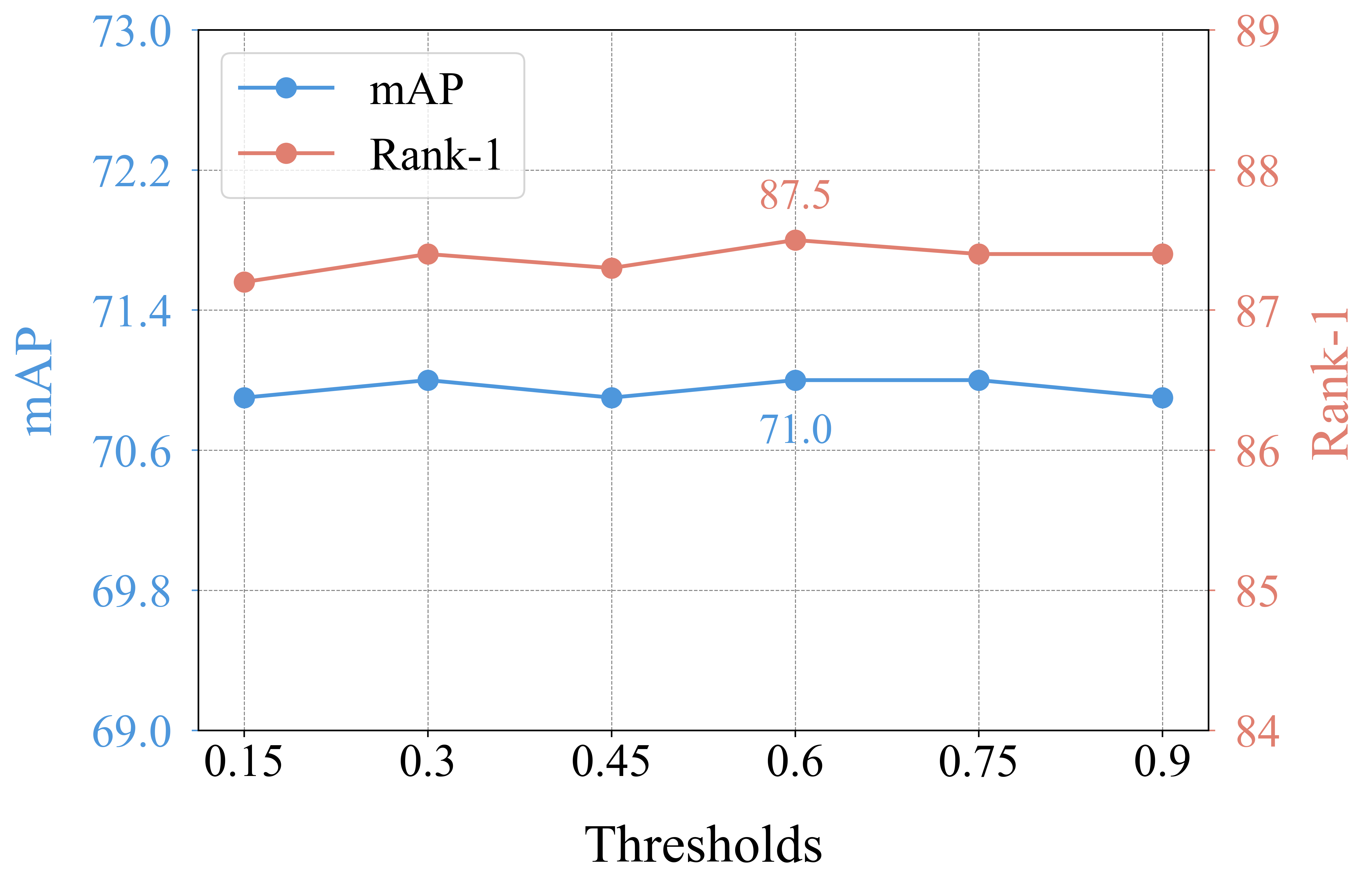}
\\
\vspace{-5pt}
\caption{\textbf{Ablation study of pose thresholds for filtering generated data.} We matched the number of training samples to ensure fairness.
}
\label{fig:supp_thres}
\vspace{-15pt}
\end{figure}

We analyzed the sensitivity of our filtering protocol to the pose confidence threshold. By varying the threshold from 0.15 to 0.9, we evaluated the CLIP-reID model on the MSMT17 dataset. Our results, presented in Figure~\ref{fig:supp_thres}, show that performance remains stable across this broad range. The mAP and Rank-1 accuracy curves are consistently high, peaking at a threshold of 0.6. This indicates that our filtering methodology is robust and not highly sensitive to the chosen threshold. We hypothesize that as long as the generated image preserves the reference identity, it provides a valuable supervision signal, even if its pose is slightly degraded.

\section{Failure Cases and Discussions}

We analyzed around 100 samples in our augmented datasets and identified two failure cases.
The first involves scenarios where the input image does not show a backpack, yet the generated image does. Although this could be considered a failure for a straightforward image-to-image generation task, it is less problematic for person re-identification because a person in the real world can appear both with and without a backpack. Generating such “hard” samples can rather actually benefit the Re-ID model by enhancing its ability to handle variations in appearance.
The second failure case arises when the input images have poor quality, such as noisy or very low-fidelity images. We believe that tackling these problems will require additional future work.

\section{Visualizing Conditions in Generative Models}
\begin{figure*}[b]
\centering
\includegraphics[width=1.0\linewidth]{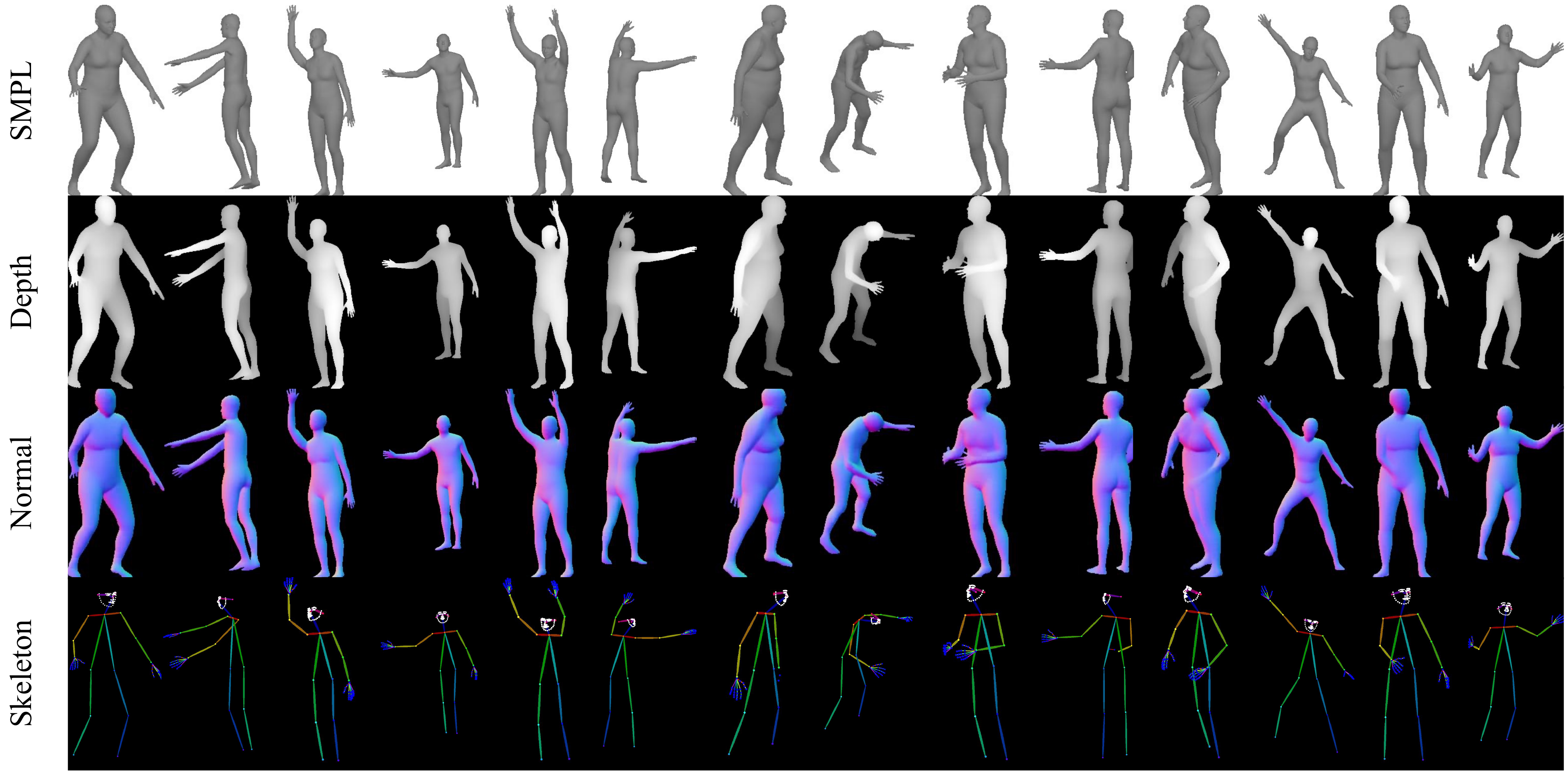}
\caption{\textbf{Example SMPL, skeleton, depth and normal maps from external dataset.} Examples of generated images from the \ours augmented datasets. The results demonstrate realistic rendering while preserving the identity of the reference images and aligning accurately with the target poses.}
\label{supl:dance_pos}
\end{figure*}
We present visualizations of conditions used in our generative model, specifically, the skeleton, depth, and normal maps, in Figure~\ref{supl:dance_pos}. With these conditions, we can control the human pose and viewpoint of the images generated by our diffusion models.

\begin{figure*}[t]
  \centering
\includegraphics[width=1.0\linewidth]{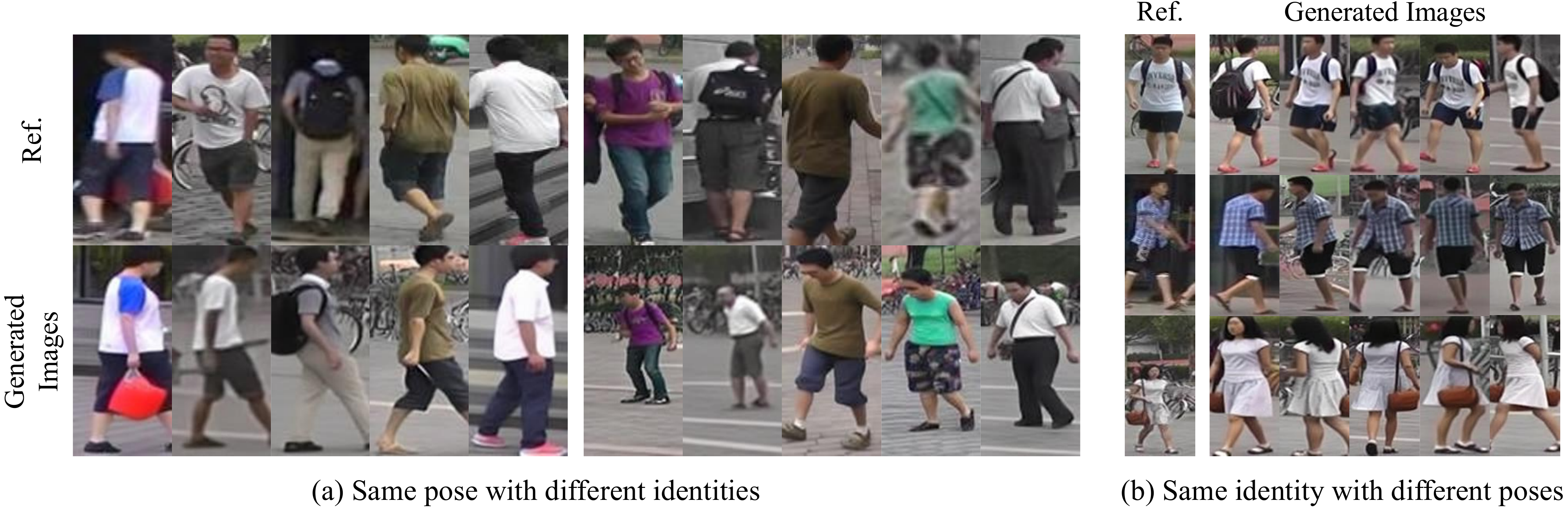}
\\
\vspace{-5pt}
\caption{\textbf{Controlled qualitative comparisons.}
(a) Different identities under the same pose and viewpoint. Identity diversity is maintained despite identical pose/viewpoint conditions. (b) Same identity across different poses and viewpoints. Identity cues are faithfully preserved while accommodating significant spatial changes.}
\label{fig:pose-id}
\vspace{-15pt}
\end{figure*}

\section{Controlled Qualitative Comparisons}
To complement the qualitative results in the main paper, we provide additional controlled comparisons in Figure~\ref{fig:pose-id}. These results are designed to explicitly evaluate identity preservation and pose/viewpoint disentanglement under two conditions. First, we show multiple poses and viewpoints for the same identity in Figure~\ref{fig:pose-id}~(a). Despite large spatial changes, our method consistently preserves identity-specific cues such as clothing and accessories. Second, we fix the same pose and viewpoint across different identities in Figure~\ref{fig:pose-id}~(b). This setting highlights that our approach can generate diverse individuals while maintaining consistency in pose and viewpoint. Together, these comparisons provide a more systematic validation that Pose-dIVE effectively separates pose/viewpoint variations from identity, enabling robust person re-identification.

\clearpage

\end{document}